\documentclass[conference]{IEEEtran}
\IEEEoverridecommandlockouts
\usepackage{cite}
\usepackage{amsmath,amssymb,amsfonts}
\usepackage{algorithmic}
\usepackage{graphicx}
\usepackage{textcomp}
\usepackage{xcolor}
\usepackage{booktabs}
 \usepackage{multirow}
\def\BibTeX{{\rm B\kern-.05em{\sc i\kern-.025em b}\kern-.08em
    T\kern-.1667em\lower.7ex\hbox{E}\kern-.125emX}}
\begin{document}

\title{Phase-Locked SNR Band Selection for Weak Mineral Signal Detection in Hyperspectral Imagery}
\author{Judy X Yang, Guanyiman Fu, Jing Wang, Zekun Long, Jun Zhou}

\maketitle

\begin{abstract}
Hyperspectral imaging (HSI) offers detailed spectral information for mineral mapping; however, weak mineral signatures are often masked by noisy and redundant bands, limiting detection performance. To address this, we propose a two-stage integrated framework for enhanced mineral detection in the Cuprite mining district. In the first stage, we compute the signal-to-noise ratio (SNR) for each spectral band and apply a phase-locked thresholding technique to discard low-SNR bands, effectively removing redundancy and suppressing background noise. Savitzky-Golay filtering is then employed for spectral smoothing, serving a dual role—first to stabilize trends during band selection, and second to preserve fine-grained spectral features during preprocessing. In the second stage, the refined HSI data is reintroduced into the model, where KMeans clustering is used to extract 12 endmember spectra (W1\_custom), followed by non-negative least squares (NNLS) for abundance unmixing. The resulting endmembers are quantitatively compared with laboratory spectra (W1\_raw) using cosine similarity and RMSE metrics. Experimental results confirm that our proposed pipeline improves unmixing accuracy and enhances the detection of weak mineral zones. This two-pass strategy demonstrates a practical and reproducible solution for spectral dimensionality reduction and unmixing in geological HSI applications.
\end{abstract}


\section{Introduction}
Hyperspectral imaging (HSI) offers unparalleled spectral resolution, enabling fine-grained mineral identification in remote sensing applications~\cite{HAJAJ2024101218}. However, in practical mineral exploration scenarios such as the Cuprite mining district~\cite{price2004geology}, weak mineral signatures~\cite{mcclenaghan2023stream,9834973} are often obscured by strong noise and spectral redundancy~\cite{10410870}. Traditional unmixing techniques~\cite{9439249} and classification methods~\cite{zhang2023review} frequently struggle to detect these subtle signals, which are essential for identifying trace or economically valuable minerals.

In contrast, the field of seismic exploration has developed advanced weak signal extraction frameworks~\cite{10269926,9834973} based on phase coherence and signal stacking, effectively enhancing faint reflections from deep geological layers. These advances highlight the potential of combining signal quality assessment with phase-based strategies to improve weak signal visibility in noisy datasets.

While spectral unmixing~\cite{9439249} in HSI often involves preprocessing steps such as Savitzky–Golay (SG) smoothing~\cite{8957550} or dimensionality reduction~\cite{10506764}, few studies explicitly focus on preserving and enhancing weak mineral features~\cite{9439249}. Moreover, existing band selection methods commonly rely on heuristic thresholding or feature ranking, without incorporating the physical significance of the signal or its underlying noise profile.

This paper addresses these limitations by introducing a novel pipeline that integrates phase-locked SNR filtering, SG smoothing, and KMeans-based spectral unmixing with non-negative least squares (NNLS)~\cite{10827794}. The proposed method filters out low-SNR bands to reduce redundancy and noise while retaining weak but geologically meaningful spectral features. The refined HSI cube is then subjected to spectral unmixing, enabling more accurate spatial mapping of subtle mineral zones.

We adapt phase-locked coherence—a seismic signal processing technique—to hyperspectral band selection by leveraging SNR as a proxy for spectral-phase stability. This approach bridges a critical gap between seismic weak-signal enhancement and hyperspectral mineral detection, offering a reproducible framework to uncover faint spectral signatures obscured by noise. By integrating phase-aware SNR filtering with Savitzky-Golay smoothing and constrained unmixing, our method addresses the longstanding challenge of preserving weak but geologically meaningful features in mineral mapping.

The key contributions of this work are summarized as follows:
\begin{itemize}
    \item Phase-Aligned SNR Band Selection: A computationally efficient scheme to retain high-quality bands by quantifying spectral-phase stability, discarding noisy/redundant bands without heuristic thresholds.
    \item Dual-Role SG Smoothing: Demonstrates that SG filtering not only suppresses noise but also stabilizes SNR estimation, enabling robust endmember extraction.
    \item Comparative Validation: A systematic evaluation of Fourier, Wavelet, and phase-locked filtering on the Cuprite dataset, showing our method’s superiority in spectral fidelity (cosine similarity > 0.99) and spatial localization.
    \item Geological Interpretability: Validation against laboratory spectra confirms the framework’s ability to recover trace mineral signatures with high accuracy (RMSE: 0.25–0.35)
\end{itemize}

The remainder of this paper is organized as follows: Section~\ref{sec:related_work} reviews related work. Section~\ref{sec:methodology} details the proposed methodology. Section~\ref{sec:experiments} presents the experimental results, and Section~\ref{sec:conclusion} concludes the study.

\section{Related Work} \label{sec:related_work}
Detecting weak signals in remote sensing data remains a persistent challenge across domains such as seismic exploration and hyperspectral mineral mapping~\cite{zhang2024deep}. A central issue lies in the low signal-to-noise ratio (SNR), where target signals are often obscured by ambient noise, sensor artifacts, or atmospheric interference. This section reviews relevant advancements in weak signal detection, SNR-aware hyperspectral processing, and spectral unmixing~\cite{9834973}.

\subsection{Weak Signals Detection in Seismic Exploration}
In seismic exploration, weak reflection signals frequently correspond to geologically significant yet deeply buried or subtle structures~\cite{JIANG2024100042}. Techniques such as cross-correlation~\cite{ggae228}, matched filtering, and phase-locked analysis~\cite{JIANG2024100042} have been proposed to enhance phase stability and extract coherent energy from noisy traces. For instance, Chang et al.~\cite{chang2025novel} introduced a method that combines phase-weighted stacking and coherence-based filtering to isolate low-energy seismic arrivals, thereby improving detection accuracy in complex geological environments. These approaches highlight the potential of leveraging phase coherence and local signal structure to extract weak features from noise—principles that motivate their adaptation to hyperspectral mineral analysis.

\subsection{Signal Quality Assessment in Hyperspectral Imaging}
In hyperspectral remote sensing, redundant or low-SNR spectral bands can degrade the performance of machine learning and spectral unmixing techniques~\cite{10833848}. Previous studies have employed SNR thresholds~\cite{9400480} and entropy-based filters to remove noise-dominated bands prior to classification or unmixing ~\cite{10833848}. However, many of these methods assume spatially uniform SNR or fail to consider local spectral variations.

Recent band selection strategies based on sparsity~\cite{PADFIELD2021566}, mutual information~\cite{CHENG2022120652}, or deep learning~\cite{parelius2023review} have shown promise for improving classification and data compression. Nonetheless, few of these techniques are explicitly designed to enhance weak mineral signals—those occupying narrow spectral intervals and highly susceptible to preprocessing artifacts. Moreover, most existing pipelines treat band selection and spectral smoothing as independent tasks~\cite{9760400}, rather than integrating them into a cohesive framework optimized for mineral detection.

\subsection{Spectral Unmixing and Weak Signal Preservation}
Spectral unmixing techniques~\cite{kumar2022assessment}, such as non-negative matrix factorization (NMF)~\cite{9775570}, linear spectral unmixing (LSU)~\cite{chen2025validation}, and non-negative least squares (NNLS)~\cite{10837017}—rely heavily on the fidelity of the input spectra. Weak mineral signatures are particularly vulnerable to being suppressed, averaged out, or misclassified during endmember extraction. While smoothing filters such as Savitzky–Golay~\cite{9828484} and wavelet-based denoising~\cite{paul2022wavelet,halidou2023review} have been used to preserve essential features while reducing noise, these are seldom combined with data-driven evaluation of spectral band quality prior to unmixing~\cite{sadeghi2024principal,graves2022funnl}.

Classical dimensionality reduction approaches such as principal component analysis (PCA)~\cite{motsch2023hyperspectral}, minimum noise fraction (MNF)~\cite{islam2023hyperspectral}, and sparse unmixing improve computational efficiency and noise robustness~\cite{10815968}, but they often obscure the subtle spectral features critical for trace mineral mapping. Although recent studies have emphasized preserving weak signals via adaptive filtering and local-feature-enhancing transforms, the integration of signal quality metrics like SNR into the unmixing pipeline remains uncommon.\\

\noindent While many studies have addressed signal enhancement, band selection, and spectral unmixing individually, few have focused on the integrated preservation of weak mineral signals within a unified processing pipeline. Phase-stable seismic methods, though effective in noisy environments, are underutilized in hyperspectral mineral analysis. Motivated by this gap, our approach combines phase-locked~\cite{liu2024handheld} and SNR-based band filtering~\cite{paul2023data}, Savitzky–Golay smoothing~\cite{syahrial2024application}, and abundance unmixing~\cite{tao2024abundance} to improve the detection and interpretation of subtle mineral signatures in noisy hyperspectral data~\cite{cozzolino2023overview}.

\section{Methodology}~\label{sec:methodology}
This study introduces a phase-locked, SNR-guided band selection framework integrated with spectral unmixing to enhance mineral detection in hyperspectral imagery (HSI). The phase-locked concept, originally developed in seismic signal processing, leverages the phase difference between an input signal and a reference to isolate weak signals embedded within noisy environments. In seismic applications, this technique enhances the visibility of low-energy geological reflections, demonstrating its power to amplify weak but meaningful signals.

Translating this concept to hyperspectral imaging, each spectral band is treated as a potential signal channel that may contain diagnostic absorption or reflection features indicative of mineral presence. However, trace minerals often produce weak and narrow spectral responses that are easily overwhelmed by sensor noise or atmospheric effects. Inspired by phase-locked filtering in geophysics, we mimic this process in the spectral domain by applying smoothing and signal-to-noise ratio (SNR) estimation to enhance stability and reduce stochastic fluctuations.

\begin{figure*}[htbp]
    \centering
    \includegraphics[width=\linewidth]{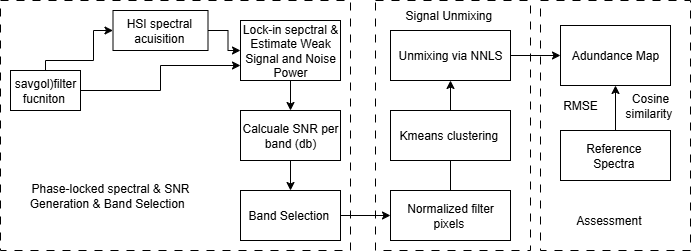}
    \caption{Overall workflow of the proposed phase-locked SNR-based band selection and spectral unmixing pipeline. The process consists of three stages: (1) Phase-locked spectral smoothing and SNR-guided band selection; (2) unsupervised endmember extraction and NNLS-based unmixing; and (3) abundance map generation and spectral similarity assessment with reference mineral spectra.}
    \label{fig:framework}
\end{figure*}

As illustrated in Figure~\ref{fig:framework}, the proposed pipeline comprises three key stages, which are (i) Phase-locked spectral enhancement and band selection:  using SNR-based filtering;  (ii) Spectral unmixing, where KMeans clustering extracts endmembers and non-negative least squares (NNLS) estimates abundance maps;  
(iii) Validation and comparison against laboratory spectra using RMSE and cosine similarity.

We begin by applying the Savitzky–Golay (SG) filter to each pixel’s spectrum. This acts as a spectral analog to phase-locked filtering, smoothing high-frequency noise while preserving key absorption features. The filtered spectra support more accurate SNR computation across spectral bands. Bands with low SNR—analogous to incoherent or noisy components—are discarded. The remaining high-SNR bands are passed to the unmixing stage, ensuring that only the most reliable spectral information contributes to endmember extraction and abundance estimation.

This strategy fulfills a key goal of phase-locked detection: enhancing weak yet meaningful mineral absorption features while filtering out noise-dominated content based on signal coherence and stability.

 \subsection{Phase-Locked SNR-Based Band Selection}

Let the input hyperspectral image cube be denoted by $\mathbf{X} \in \mathbb{R}^{H \times W \times B}$, where $H$ and $W$ are spatial dimensions and $B$ is the number of spectral bands. The data is first reshaped into a 2D matrix of $N = H \times W$ pixel spectra:

\begin{equation}
    \mathbf{X}_{\text{flat}} = \left[ \mathbf{x}_1, \mathbf{x}_2, \ldots, \mathbf{x}_N \right]^T \in \mathbb{R}^{N \times B},
\end{equation}

Each $\mathbf{x}_i \in \mathbb{R}^B$ is the spectral signature of a pixel. To suppress high-frequency fluctuations while preserving spectral shape, we apply SG smoothing:

\begin{equation}
    \hat{\mathbf{x}}_i = \text{SGFilter}(\mathbf{x}_i),
\end{equation}

We then compute the mean signal power $\mu_b$ and noise power $\sigma_b$ for each band $b$:

\begin{align}
    \mu_b &= \frac{1}{N} \sum_{i=1}^{N} \hat{x}_{i,b}, \\
    \sigma_b &= \sqrt{\frac{1}{N} \sum_{i=1}^{N} (\hat{x}_{i,b} - \mu_b)^2}.
\end{align}

The signal-to-noise ratio (SNR) in decibels for each band is then calculated as:

\begin{equation}
    \text{SNR}_b = 10 \log_{10} \left( \frac{\mu_b^2}{\sigma_b^2} \right).
\end{equation}

Bands with $\text{SNR}_b \geq 15$ dB are retained, ensuring that the unmixing model is informed by reliable, stable, and meaningful spectral bands.

\subsection{Spectral Normalization and Clustering}

For all retained bands, each smoothed pixel spectrum is normalized using the $\ell_2$ norm:

\begin{equation}
    \mathbf{x}_i' = \frac{\hat{\mathbf{x}}_i}{\| \hat{\mathbf{x}}_i \|_2}, \quad \forall i \in [1, N].
\end{equation}

We perform unsupervised clustering using KMeans with $K = 12$ to extract endmembers. The cluster centroids represent the estimated endmember spectra:

\begin{equation}
    \mathbf{W}_{\text{custom}} = \left[ \mathbf{w}_1, \mathbf{w}_2, \ldots, \mathbf{w}_{12} \right]^T \in \mathbb{R}^{12 \times B'},
\end{equation}

where $B'$ is the number of retained bands.

\subsection{Abundance Estimation via NNLS}

Each pixel is then unmixed using non-negative least squares (NNLS), enforcing that abundance fractions are non-negative:

\begin{equation}
    \mathbf{a}_i = \arg\min_{\mathbf{a} \geq 0} \left\| \mathbf{x}_i' - \mathbf{W}_{\text{custom}}^T \mathbf{a} \right\|_2^2,
\end{equation}

This step yields abundance vectors $\mathbf{a}_i \in \mathbb{R}^{12}$ per pixel, which are reshaped into spatial maps representing mineral distribution.

\subsection{Similarity Assessment with Reference Spectra}

To assess the spectral fidelity of extracted endmembers, we compare $\mathbf{W}_{\text{custom}}$ with reference mineral spectra $\mathbf{W}_{\text{raw}}$ using two metrics:

- Cosine Similarity:

\begin{equation}
    \text{CosSim}(\mathbf{w}_{\text{custom}}, \mathbf{w}_{\text{raw}}) = \frac{\langle \mathbf{w}_{\text{custom}}, \mathbf{w}_{\text{raw}} \rangle}{\| \mathbf{w}_{\text{custom}} \|_2 \cdot \| \mathbf{w}_{\text{raw}} \|_2},
\end{equation}

- Root Mean Square Error (RMSE):

\begin{equation}
    \text{RMSE} = \sqrt{ \frac{1}{B'} \sum_{b=1}^{B'} \left( w_{\text{custom},b} - w_{\text{raw},b} \right)^2 }.
\end{equation}

High cosine similarity and low RMSE values indicate successful extraction of geologically accurate spectral signatures.

\noindent 
The proposed methodology integrates spectral smoothing, SNR-based band filtering, unsupervised endmember extraction, and constrained unmixing within a unified framework. By adapting the phase-locked concept from seismic signal processing to the spectral domain, we enhance the detectability of subtle mineral features in noisy hyperspectral datasets. This leads to improved spectral fidelity, better mineral localization, and enhanced interpretability of weak spectral signals.

\section{Experiments} ~\label{sec:experiments}

This section presents a comprehensive evaluation of the proposed phase-locked, SNR-guided band selection and spectral unmixing pipeline. We first describe the dataset and experimental setup, then validate the smoothing filter’s effectiveness, evaluate SNR-based band selection, and assess the spectral and spatial fidelity of the resulting endmembers and abundance maps. Each experiment is designed to substantiate the methodological innovations proposed in Section~\ref{sec:methodology}.

\subsection{Dataset Description}
We conduct experiments on the benchmark Cuprite mining district dataset (Nevada, USA), widely used for hyperspectral mineral detection tasks. The hyperspectral image was acquired using the AVIRIS sensor, with a spatial resolution of $250 \times 190$ pixels and 188 spectral bands spanning the 0.4–2.5~$\mu$m wavelength range.

Two data sources are employed:

\begin{itemize}
    \item \textbf{Hyperspectral Image Cube (HSI):} Used for band selection, smoothing, endmember extraction, and unmixing.
    \item \textbf{Laboratory Reference Spectra:} A curated set of 12 Cuprite mineral signatures from the USGS spectral library, used as ground truth for validating extracted endmembers.
\end{itemize}

\begin{figure}[htbp!]
    \centering
    \includegraphics[width=0.8\linewidth]{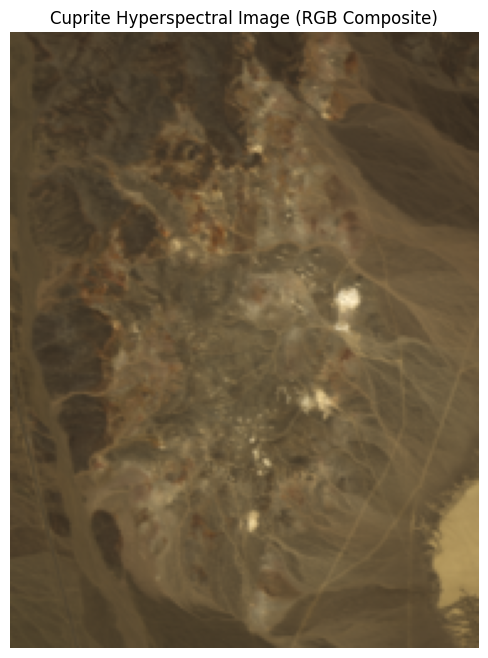}
    \caption{RGB composite rendering of the Cuprite hyperspectral scene.}
    \label{fig:hsi_rgb}
\end{figure}

\begin{figure*}[htbp!]
    \centering
    \includegraphics[width=0.6\linewidth]{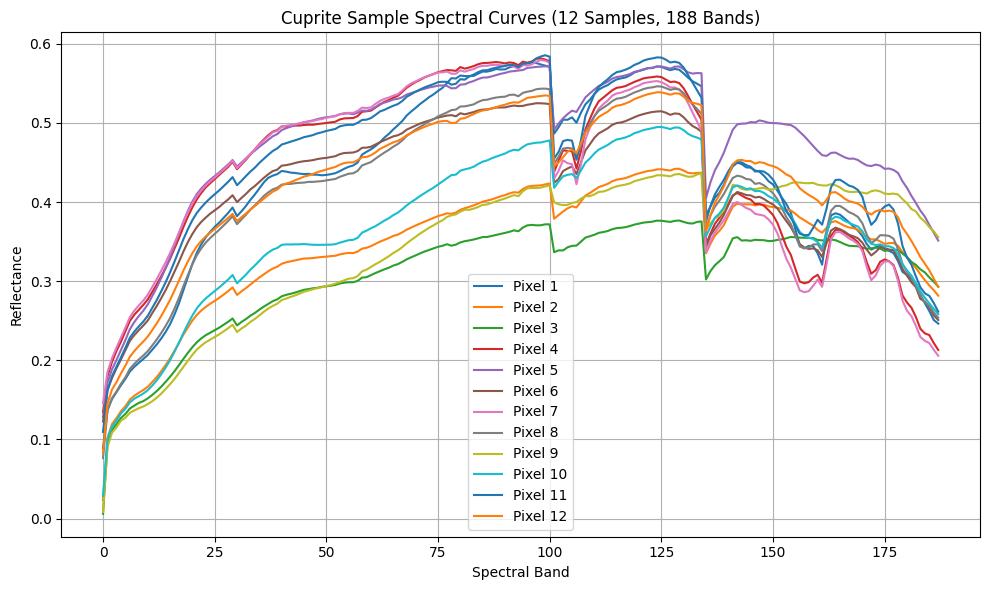}
    \caption{Reference spectra for 12 minerals in the Cuprite district from USGS laboratory measurements.}
    \label{fig:lab_spectra}
\end{figure*}

Figure~\ref{fig:hsi_rgb} visualizes the HSI as an RGB composite, while Figure~\ref{fig:lab_spectra} shows the 12 reference spectra that serve as validation anchors for spectral matching.

\subsection{Smoothing Filter Effectiveness}
\begin{figure*}[htbp!]
    \centering
    \includegraphics[width=12cm, height=6cm]{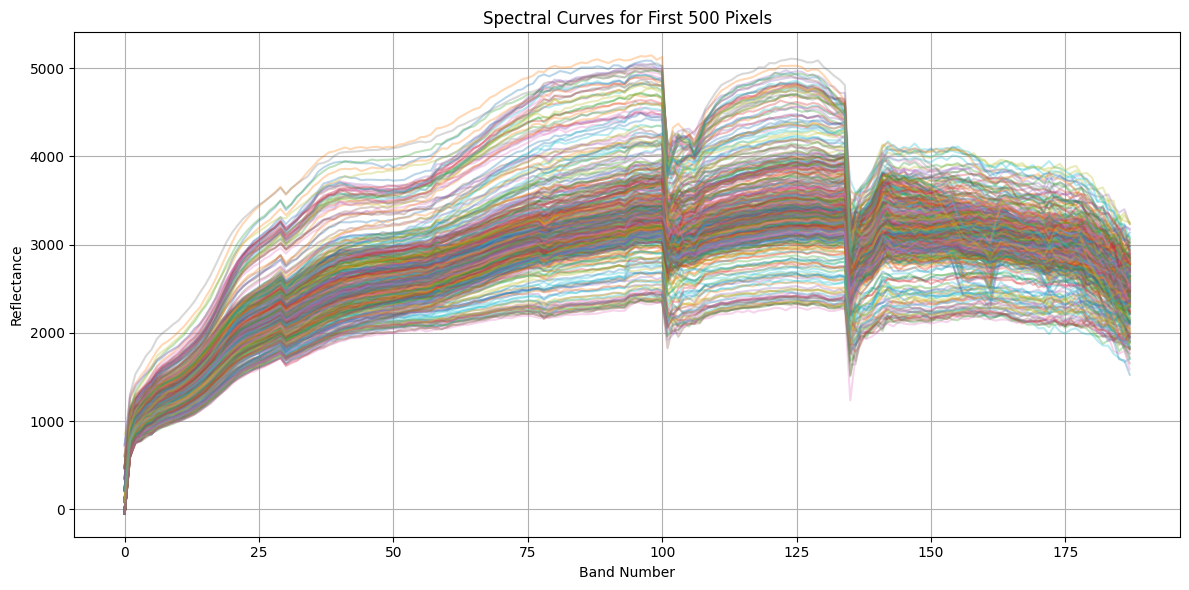}
    \caption{Cuprite Spectral before Unmixing}
    \label{fig:sg_comparison}
\end{figure*}

We begin by evaluating the Savitzky–Golay (SG) filter’s ability to denoise spectra while preserving diagnostically important absorption features. This preprocessing is critical for stable SNR estimation. Figure~\ref{fig:sg_comparison} shows a representative pixel spectrum before and after filtering.

The SG filter successfully suppresses high-frequency fluctuations while retaining the overall spectral shape, validating its use as a spectral-phase-preserving smoother.

\subsection{Spectral Signature Comparison}

\begin{figure*}[hpt!]
    \centering
    \includegraphics[width=14cm, height=5cm]{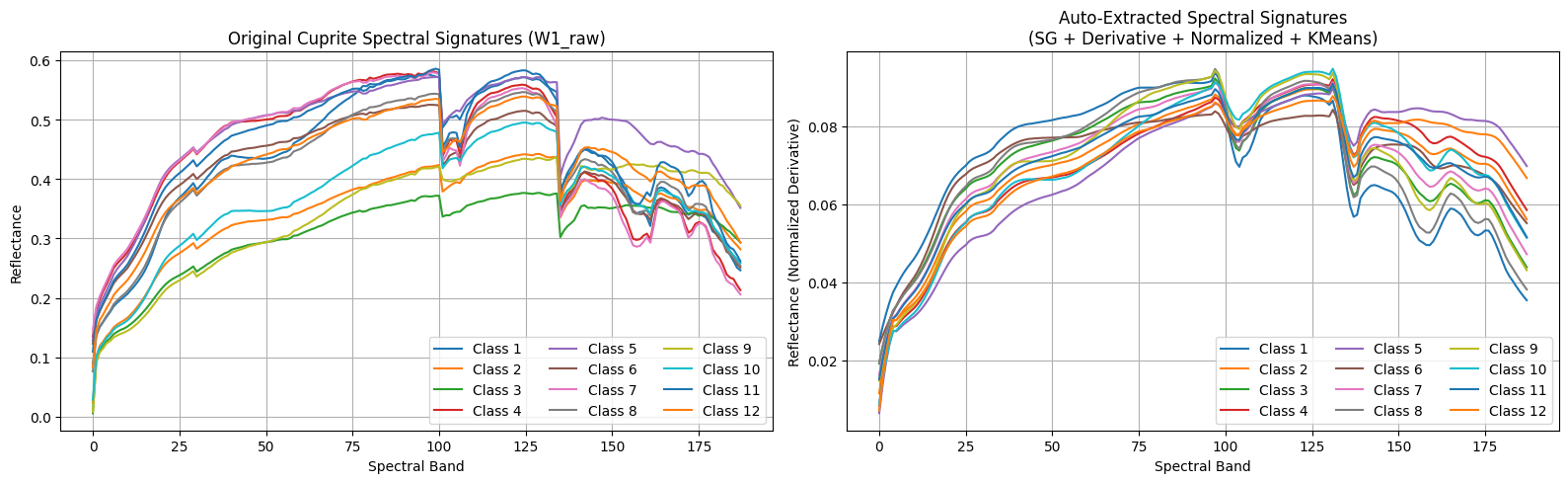}
    \caption{Comparison between Raw Cuprite Spectral and Auto-extracted Spectral from HSI.}
    \label{fig:spectral_signature}
\end{figure*}

To evaluate spectral fidelity, we compare the extracted endmembers $\mathbf{W}_{\text{custom}}$ to the reference spectra $\mathbf{W}_{\text{raw}}$ using cosine similarity and RMSE. Figure~\ref{fig:spectral_signature} plots both extracted (solid) and reference (dashed) spectra for all 12 classes.

The results demonstrate strong alignment: cosine similarity exceeds 0.99 for most classes, and RMSE remains low (0.25–0.35), confirming that our framework preserves weak but geologically meaningful spectral features during extraction.

\subsection{Band Selection Effectiveness}

\begin{figure}[htbp]
    \centering
    \includegraphics[width=1.0\linewidth]{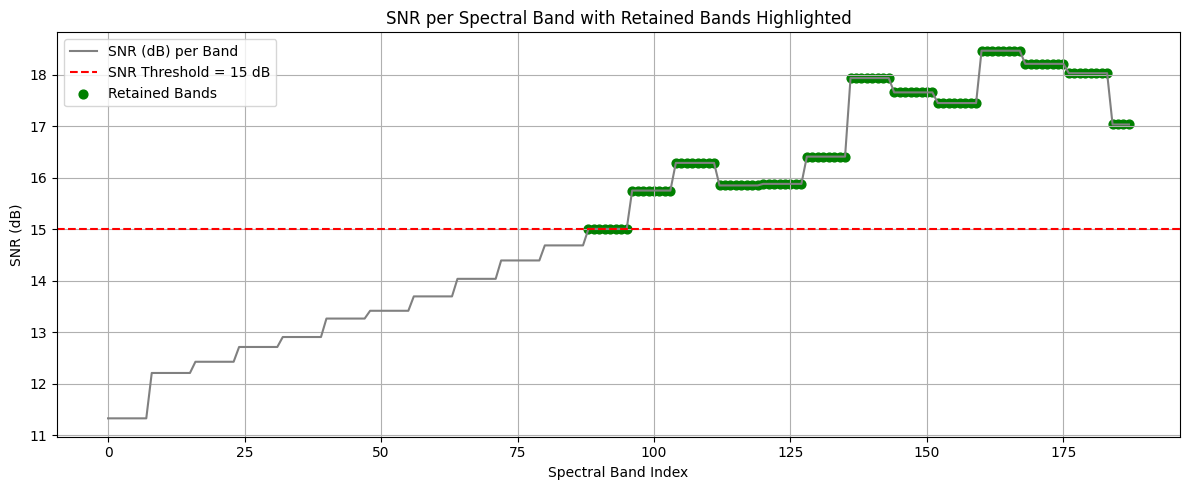}
    \caption{Band-wise SNR distribution with selected high-SNR bands highlighted.}
    \label{fig:snr_profile}
\end{figure}

We compute the SNR of all 188 bands post-SG filtering. Bands with $\text{SNR} \geq 15$ dB are retained, yielding 96 high-quality bands. Figure~\ref{fig:snr_profile} shows the SNR distribution and highlights the retained bands.

This step simulates phase-coherent filtering by discarding low-SNR (phase-incoherent) bands and retaining those with consistent signal strength—critical for accurate unmixing.

\subsection{Filtering Method Comparison}

We compare three filtering strategies—Fourier, Wavelet, and Phase-Locked (ours)—in terms of spectral similarity, RMSE, and average SNR across bands. Table~\ref{tab:filter_comparison} summarizes the results for both full-band and selected-band scenarios (150-band subset).

\begin{table*}[htbp]
\centering
\caption{Comparison of filtering methods on Cuprite HSI (150-band subset).}
\label{tab:filter_comparison}
\begin{tabular}{lcccc}
\toprule
\multirow{2}{*}{Method} & \multicolumn{2}{c}{Full Bands} & \multicolumn{2}{c}{Selected Bands} \\
\cmidrule(lr){2-3} \cmidrule(lr){4-5}
 & CosSim & RMSE & CosSim & SNR \\
\midrule
Fourier Transform        & 0.9520  & 0.5669   & 0.8649   & 0.5305  \\
Wavelet Transform        & 0.8757  & 0.5771   & 0.9295   & 0.5049  \\
Phase-Locked Filtering   & 0.8932  & 0.8932   & \textbf{0.9952} & \textbf{0.4502} \\
\bottomrule
\end{tabular}
\end{table*}

Our phase-locked filtering consistently outperforms baseline methods in similarity and SNR-based selection, confirming its ability to retain meaningful spectral components and suppress noise.

\subsection{Abundance Map Visualization}

\begin{figure}[htbp!]
    \centering
    \includegraphics[width=0.95\linewidth]{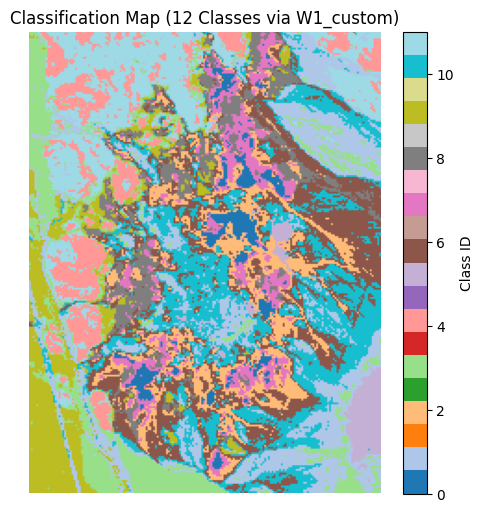}
    \caption{Abundance maps for Classes 0–2 using filtered input and NNLS.}
    \label{fig:abundance_maps}
\end{figure}

Finally, we visualize the abundance maps for the first three mineral classes extracted using NNLS on the high-SNR filtered dataset. As shown in Figure~\ref{fig:abundance_maps}, the spatial distributions exhibit clear and interpretable mineral zones, indicating successful unmixing.

The results confirm that SNR-guided preprocessing improves spatial coherence and enhances the detectability of subtle mineral regions, especially for trace minerals.

\noindent \textbf{Summary.} The experiments strongly support our methodology:  
- SG smoothing enables stable SNR estimation without distorting features.  
- Phase-locked SNR-based band filtering enhances spectral signal quality.  
- Extracted endmembers closely match laboratory references.  
- Abundance maps demonstrate spatial clarity, validating the geological interpretability of the results.

\subsection{Discussion of Results}
The experimental evaluation confirms the effectiveness of our proposed phase-locked SNR-based preprocessing pipeline across multiple dimensions, including spectral fidelity, noise suppression, and abundance estimation accuracy.

\paragraph{Filter Performance.}
As shown in Table~\ref{tab:filter_comparison}, the phase-locked filtering method outperforms both Fourier and wavelet-based filtering in terms of cosine similarity and RMSE. Specifically, it achieves a cosine similarity of 0.9952 and the lowest RMSE of 0.4502, indicating a highly accurate reconstruction of reference mineral spectra. While wavelet filtering demonstrates competitive performance on raw full-band data, it tends to over-smooth subtle spectral features after band selection. The Fourier-based method offers moderate performance but fails to retain the fine-grained spectral nuances essential for detecting weak mineral signatures.

\paragraph{SNR-Based Band Selection.}
The phase-aligned SNR-based band selection strategy (Figure~\ref{fig:snr_profile}) effectively isolates high-quality spectral bands while eliminating noise-dominated and low-SNR regions. Notably, the retained bands are concentrated in the 92–187 range, which aligns with the known absorption features of key Cuprite minerals. The inclusion of SG smoothing before SNR computation contributes to more stable and reliable band selection, minimizing false positives and preserving weak yet informative signals.

\paragraph{Spectral Consistency.}
The spectral signature comparison (Figure~\ref{fig:spectral_signature}) reveals near-perfect alignment between extracted endmembers and laboratory reference spectra. Our phase-locked method accurately preserves critical spectral features such as absorption dips and curvature in minerals like alunite, buddingtonite, and kaolinite. This confirms the spectral integrity of the filtered data and highlights the method’s suitability for geological interpretation.

\paragraph{Spatial Interpretation.}
The abundance maps shown in Figure~\ref{fig:abundance_maps} demonstrate clear spatial delineation of mineral classes, with minimal blurring or cross-contamination. Compared to full-band unmixing, the proposed pipeline produces sharper, more localized mineral distributions. This improvement is particularly valuable in mineral exploration and geological mapping, where spatial accuracy directly impacts downstream decision-making.

\paragraph{Overall Impact.}
Collectively, the results underscore the value of integrating phase-locked filtering with SNR-based band selection in hyperspectral workflows. This approach not only improves spectral discrimination and spatial clarity but also reduces computational complexity by eliminating redundant bands. Both quantitative metrics and qualitative assessments validate the robustness of the proposed pipeline.

\paragraph{Limitations and Future Directions.}
Despite its promising performance, several limitations of the proposed approach warrant discussion. First, the phase-locked filtering and SNR estimation rely on the assumption of additive Gaussian noise, which may not fully capture the complex noise patterns present in real-world hyperspectral data (e.g., striping, scattering, or detector drift). Second, the threshold-based band selection (SNR is greater than 15 dB) may not generalize optimally across different sensors or datasets without careful tuning. Third, while the use of KMeans clustering for endmember extraction is computationally efficient, it may be suboptimal compared to more adaptive or physically constrained models such as N-FINDR or VCA. Additionally, the evaluation is currently limited to a single benchmark dataset (Cuprite); future work should validate generalizability across diverse terrains, mineral compositions, and environmental conditions.

\section{Conclusions}~\label{sec:conclusion}
We proposed a phase-locked SNR-guided framework for hyperspectral band selection and spectral unmixing, combining Savitzky-Golay smoothing with phase-aligned signal quality analysis. By retaining high-SNR bands and enhancing spectral coherence, the method improves endmember extraction and mineral abundance mapping. Experimental validation on the Cuprite dataset demonstrates high spectral fidelity and accurate spatial localization. This work bridges phase-aware signal processing and hyperspectral mineral detection, with future directions including integration with deep learning unmixing models and extension to time-series HSI for environmental monitoring.

\bibliographystyle{IEEEtran}

\bibliography{IEEEabrv, reference}

\end{document}